\documentclass[letterpaper]{article} 
\usepackage{aaai2026}  
\usepackage{times}  
\usepackage{helvet}  
\usepackage{courier}  
\usepackage[hyphens]{url}  
\usepackage{graphicx} 
\usepackage{natbib}  
\usepackage{caption} 
\usepackage{algorithm}
\usepackage{algorithmic}

\usepackage{booktabs}
\usepackage{array} 

\usepackage{multirow}
\usepackage{tikz}
\usepackage{amssymb}
\usepackage{xspace}
\usepackage{tabularx}
\usepackage{enumitem}
\usepackage[T1]{fontenc}
\usepackage{color}
\definecolor{myred}{rgb}{.8,.0,.0}
\definecolor{myblue}{rgb}{.0,.0,.8}

\def\etal{\textit{et al.}\xspace}
\newcommand{\DarZaj}{Dariusz Zaj\k{a}c}
\newcommand{\JimSan}{Jim\'{e}nez-S\'{a}nchez}
\newcommand{\checkbox}{\raisebox{0.5pt}{\large $\square$}}

\usepackage[most]{tcolorbox}
\definecolor{lightpurple}{rgb}{0.75, 0.58, 1.0}
\definecolor{mypurple}{rgb}{.4,.0,.8}
\newcommand{\disclaimer}[1]{\textcolor{mypurple}{\textbf{#1}}}

\newtcolorbox{disclaimerbox}{
  colframe=lightpurple!50,   
  colback=lightpurple!5,     
  boxrule=0.6mm,            
  width=\textwidth          
}

\usepackage{longtable}

\usepackage{newfloat}
\usepackage{listings}
\DeclareCaptionStyle{ruled}{labelfont=normalfont,labelsep=colon,strut=off} 
\floatstyle{ruled}
\newfloat{listing}{tb}{lst}{}
\floatname{listing}{Listing}
\nocopyright

\makeatletter
\g@addto@macro\@maketitle{
	\vspace{-1.2em}  
    \begin{center}
    \begin{disclaimerbox}
  	\disclaimer{Disclaimer: this is a working paper, and represents research in progress. We welcome contributions from the community, for comments or questions please email us at yucheng.lu@uantwerpen.be, vech@itu.dk, and amelia.jimenez@ub.edu.}
\end{disclaimerbox}
    \end{center}
    \vspace{-0.5em}
}
\makeatother

\title{Intuitions of Machine Learning Researchers about \\ Transfer Learning for Medical Image Classification}
\author{
	Yucheng Lu\textsuperscript{\rm 1,3\textdagger}, 
	Hubert \DarZaj\textsuperscript{\rm 2}, 
	Veronika Cheplygina\textsuperscript{\rm 3}, 
	Amelia \JimSan\textsuperscript{\rm 3,4\textdagger}
	}
\affiliations{ 
    \textsuperscript{\rm 1}University of Antwerp, Belgium\\
    \textsuperscript{\rm 2}University of Copenhagen, Denmark\\
    \textsuperscript{\rm 3}IT University of Copenhagen, Denmark\\
    \textsuperscript{\rm 4}Universitat de Barcelona, Spain\\[0.5em]

}

\begin{document}
\pagestyle{plain} 

\maketitle

\begingroup
\renewcommand{\thefootnote}{\textdagger}
\footnotetext{This work was carried out while YL and AJS were at the IT University of Copenhagen. YL is currently affiliated with the University of Antwerp, Belgium, and AJS with the University of Barcelona, Spain.}
\endgroup

\begin{abstract}

Transfer learning is crucial for medical imaging, yet the selection of source datasets often relies on researchers' intuition rather than systematic principles, which can impact the generalizability of algorithms and, thus, patient outcomes. This study investigates these decisions through a task-based survey with machine learning practitioners. Unlike prior work that benchmarks models and experimental setups, we take a human-computer interaction (HCI) perspective on how practitioners select source datasets. Our findings indicate that choices are task-dependent and influenced by community practices, dataset properties, and computational (data embedding), or perceived visual or semantic similarity. However, similarity ratings and expected performance are not always aligned, challenging a traditional ``more similar is better'' view. Moreover, ethical and fairness considerations remain largely absent from source dataset sections. Participants often used ambiguous terminology, which suggests a need for clearer definitions and tools to make them explicit and usable. By clarifying these heuristics and introducing a conceptual framework of transfer learning factors, this work provides practical insights for more systematic source selection in transfer learning.
\end{abstract}


\section{Introduction}
\begin{figure*}[htb]
    \centering
    \includegraphics[width=\textwidth]{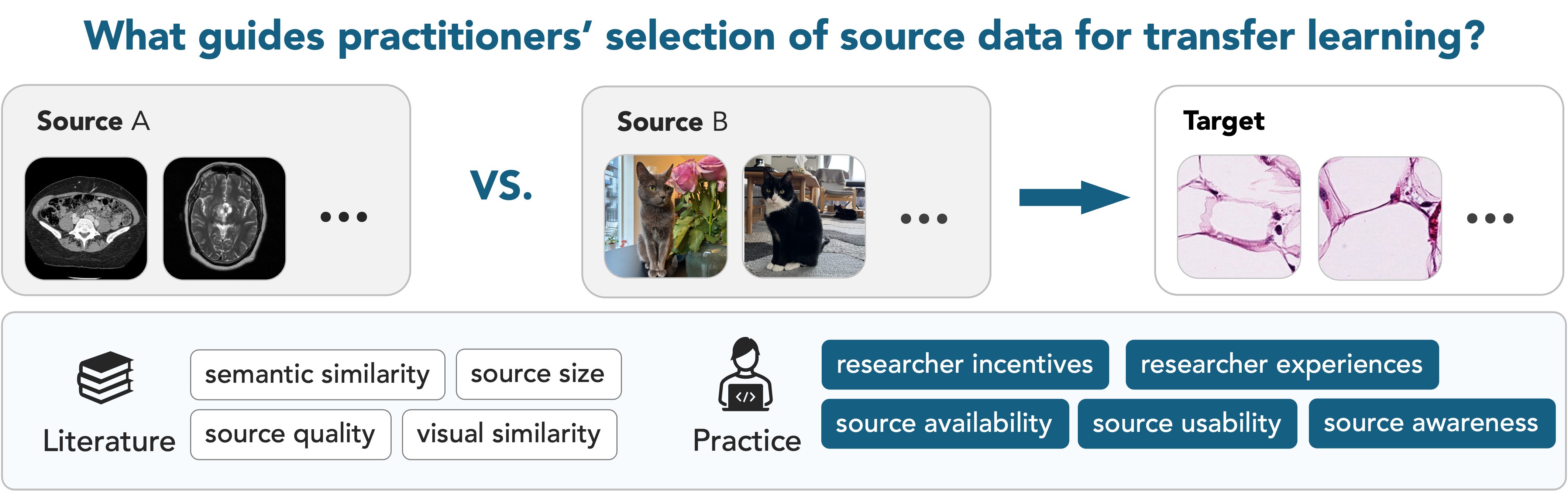}
    \caption{Transfer learning source selection in the context of medical image classification is influenced by characteristics such as visual and semantic similarity, dataset size, and task complexity, shaping model adaptation and performance. \textit{Source A} and \textit{Target} are semantically similar but visually dissimilar, as both are medical images but they do not share visual characteristics. In contrast, \textit{Source B} and \textit{Target} are visually similar but semantically dissimilar.}
    \label{fig:abstract}
\end{figure*}
Deep learning (DL) has become a cornerstone of modern machine learning (ML), driving advances in areas ranging from image recognition to natural language processing \cite{shinde_review_2018}. These developments are often fueled by access to massive, general-purpose datasets. Yet, when DL techniques are applied to specialized domains such as medical imaging, the availability of high-quality, task-specific training data becomes a significant bottleneck \cite{janiesch_machine_2021}. First, what constitutes high-quality data is context-dependent \cite{mohammed_data_2024, zajkac2023ground}. Second, the best attempt at striving for it requires vast human resources, such as the time of specialized clinicians \cite{li2023systematic, zajkac2023ground}. To address this challenge, researchers are increasingly turning to transfer learning -- a strategy that adapts models trained on large, general \emph{source datasets} (e.g., from computer vision) to perform well on domain-specific tasks (e.g., medical imaging) using much smaller, curated \emph{target datasets} \cite{pouyanfar_survey_2018, cheplygina2019cats}.

Numerous studies have explored the concrete applications of transfer learning, including various criteria related to the \emph{source} and \emph{target datasets} that influence its success, such as size~\cite{raghu2019transfusion, mensink2021factors}, task complexity~\cite{ribeiro2017exploring}, semantic similarity~\cite{chen2019med3d}, visual similarity~\cite{shi2018learning}, and feature space similarity~\cite{juodelyte2024dataset}. While insightful, these studies often focus on a limited number of factors, making it challenging to apply these insights to other projects. Particularly, there is little consensus on how researchers choose \emph{source datasets} and which factors are considered important for effective transfer learning. As a result, experienced ML engineers often rely on intuition when deciding on the best parameters.

The responsible AI and human–computer interaction (HCI) communities have a long-standing interest in examining expert work to better understand decision-making and to inform the design of systems that are grounded in real-world practice \cite{schmidt_trouble_2012, alvarado_garcia_emerging_2025, wang_whose_2022}. This includes efforts to surface and theorize the tacit knowledge and intuition that guide the work of data science and ML practitioners. For instance, Muller~\etal~\cite{muller_designing_2021} explored how data workers navigate uncertainty and make situated decisions in ML workflows, often drawing on informal practices and experiential knowledge. Building on this, Cha~\etal~\cite{cha_unlocking_2023} investigated how ML practitioners rely on tacit understandings when constructing datasets, showing that data creation is deeply contextual, shaped by the individuals involved and tightly coupled with the models.

Extending these traditions of explicating tacit knowledge, this study addresses the following core research question: In the high-stakes context of medical imaging, what tacit heuristics, social dynamics, and ethical considerations guide ML practitioners when selecting source datasets? We focus on medical imaging because opaque implementation decisions, such as the intuitive selection of a pre-training source, can lead to biased models or misdiagnoses. By examining expert choice intuition, we aim to make these consequential decisions explicit. While empirically focused on medical imaging, our framework incorporates general ML literature because medical projects frequently adopt general computer vision baselines. To investigate this, we engaged $N=15$ practitioners in a mixed-methods survey involving their recent projects and two distinct control case studies (see Fig.~\ref{fig:abstract}) to contextualize their selection intuitions.

\begin{enumerate}
    \item We reveal that \emph{source datasets} selection is not only a rational process driven by technical aspects like domain alignment but also a result of social dynamics influenced by established baselines, community practices, model availability, and reviewers' expectations.

    \item We discuss how ethical and fairness considerations in transfer learning are acknowledged but deprioritized in relation to other social and technical criteria, due to the lack of incentive structures and evaluation frameworks that would make it actionable.

    \item We map the criteria considered by researchers when adopting transfer learning into a conceptual framework, providing a consistent vocabulary for future empirical research on data selection in ML.

    \item Our findings confirm the importance of embedding similarity alongside semantic and visual similarity, understood as texture and structure. However, the misalignment between similarity ratings and expected performance challenges the ``more similar is better'' approach. We also identify frequent yet vague use of concepts like ``good image quality'' and ``domain gap'', underscoring the need for precise operational definitions and tools to make these notions actionable.
\end{enumerate}

\section{Related work}
\subsection{Many faces of tacit knowledge in ML work}

HCI researchers conceptualize the overlooked work and knowledge underpinning ML pipelines, with a recent emphasis on data work. Originally studied by Bowker and Star \cite{bowker_sorting_2000}, such work is vital as contemporary ML systems depend on vast, curated datasets \cite{wang_whose_2022}. Miceli~\etal~\cite{miceli_data-production_2022} found that the \emph{truth} encoded in datasets is not a neutral representation of reality but a product of situated labor mediated by socioeconomic and organizational constraints. Similarly, Muller~\etal~\cite{muller_designing_2021} showed how practitioners use tacit knowledge during data labeling to navigate quality issues. Their work calls for a deeper theorization of tacit knowledge in ML practice, a direction we build upon in this paper.


However, data work in ML extends beyond annotation to encompass broad activities like data preparation and transformation \cite{muller2019data}. Alvarado Garcia~\etal~\cite{alvarado_garcia_emerging_2025} examined how LLM development shapes practitioners' handling of uncertainty and data practices, highlighting new opportunities for HCI to address generative AI's ethical challenges. Complementing this, Cha~\etal~\cite{cha_unlocking_2023} investigated the role of tacit knowledge in dataset creation. They revealed \emph{what} forms of tacit knowledge are mobilized and \emph{why} such knowledge is indispensable, as data remains context-dependent, inseparable from human workers, and tied to specific models. This work advocates for moving from ad-hoc practices toward systematic ways of articulating tacit knowledge within ML pipelines.


Further, ML model development is guided as much by assumptions and intuition as by measurable evidence. Investigations into this implicit knowledge include work by Cabrera~\etal~\cite{cabrera_what_2023}, who studied the mental models of ML engineers regarding model learning. They developed a tool to support the understanding of model behaviors, which effectively explicates and enhances the tacit assumptions that shape model choice.


Finally, particularly relevant is transfer learning, which adapts models from \emph{source datasets} to \emph{target datasets} for domain-specific tasks. This strategy is increasingly studied in HCI, such as Zeng~\etal~\cite{zeng_intenttuner_2024} who developed IntentTuner to integrate human intentions into fine-tuning workflows through structured data processing strategies. Conversely, Mishra~\etal~\cite{mishra_designing_2021} explored how non-experts perceive transfer learning, finding that even domain experts face progress hurdles due to misunderstandings of the underlying learning process. These studies illustrate efforts to conceptualize the tacit knowledge of data workers and translate them into concrete guidance for ML pipelines.


Importantly, tacit knowledge is not a mysterious insight but a critical heuristic for decision-making under uncertainty, encompassing cooperative and socially situated practices \cite{schmidt_trouble_2012}. In transfer learning, the search space for an ``optimal'' \textit{source dataset} is prohibitively vast across numerous datasets and architectures. Because relying solely on exhaustive empirical evidence is computationally and temporally infeasible, practitioners employ \emph{heuristic filters} that rely on internalized community norms, established practices, or past experiences rather than technical optimization. This process informs selection behaviors that prioritize efficiency and social legitimacy over pure technical ``best fit''. While model training and data creation are well-studied, transfer learning in data-scarce domains like medical imaging remains guided by this intuition. How ML researchers evaluate and select such data remains largely unexplored, leaving a key aspect of real-world ML practice invisible. Our human-centered study brings to the surface hidden and under-studied aspects of decision-making in transfer learning, advancing transparency in foundational research.

\subsection{Transfer learning in medical imaging}
Transfer learning has become a key approach in medical imaging, addressing the challenge of limited dataset sizes in this domain \cite{cheplygina2019cats,litjens2017survey,kim2022transfer}. In short, a model is first trained on the \emph{source dataset}, and then fine-tuned on the \emph{target dataset}. In this process, there are several factors influencing the results, such as the datasets, model architectures, evaluation metrics, and fine-tuning strategies, which makes it challenging to compare results or draw general conclusions.

In practice, transfer learning approaches are often reduced to testing arbitrary fine-tuning configurations without clear justifications \cite{hemelings2020accurate}, or not describing them completely \cite{valkonen2019cytokeratin,han2018deep}. This reflects a broader pattern observed in the ML community, where development of novel algorithms often takes precedence over the critical examination of datasets, which are frequently treated as neutral or objective benchmarks \cite{birhane2021values,sambasivan2021everyone}. 

Regarding \emph{source data} for pretraining, ImageNet-1K \cite{deng2009imagenet} remains a standard in medical imaging due to its scale and the reduced workload offered by available models. However, medical images rely on subtle local textures rather than the prominent global structures of natural images. As transfer learning is most effective when source and target domains share similar distributions \cite{pan2009survey}, ImageNet-1K may be unsuitable for medical classification, especially in low-data regimes \cite{raghu2019transfusion}. To address this, domain-specific datasets like RadImageNet \cite{mei2022radimagenet}, Med3D \cite{chen2019med3d}, and VOCO \cite{wu2024large} have been developed to better reflect medical image characteristics. Despite their potential, these sources are not yet widely adopted because some, like RadImageNet, are only available upon request.

When selecting a \emph{source dataset} for transfer learning, research points to several other considerations, alongside visual similarity. Two commonly cited factors are: (i) a sufficient amount of data to train a model from scratch, and semantic alignment between the pretraining and target domains, specifically, whether the \emph{source dataset}  comprises natural or medical images. Additional characteristics have also been identified as influential in cross-domain transferability, such as the dimensionality of the images (2D or 3D), or number of classes, see \cite{cheplygina2019cats} for examples of each in medical imaging. 


Despite conceptualization efforts, terms like representativeness and diversity are often invoked without clear definitions when motivating \emph{source datasets} selection or evaluating transfer learning outcomes. This ambiguity hinders ML model reliability and reflects broader issues across the field. To address these challenges, Clemmensen~\etal{}~\cite{clemmensen2022data} reviewed various interpretations of \emph{data representativity} regarding valid inference, while Zhao~\etal{}~\cite{zhao2024position} provided recommendations for conceptualizing and evaluating dataset \textit{diversity}.

However, we still lack a grounded understanding of how practitioners themselves interpret and apply such notions in practice. In particular, the selection of the \emph{source dataset} for transfer learning and the relevance of its dimensions are often guided by intuition rather than a systematic framework. This gap highlights the need for empirical investigation into the tacit criteria that influence dataset choice in transfer learning. 

\section{Conceptualization of transfer learning factors}
\label{sec:factors}



Many studies both within and outside medical imaging examine factors contributing to the success of transfer learning, or transferability, which depends on the \emph{source dataset}, \emph{target dataset}, model architecture, and fine-tuning strategy. While not an exhaustive review, we describe key factors in Table~\ref{tab:categories} (with additional examples in Appendix~\ref{appendix:factors}), noting that these elements are often interdependent, such as a smaller domain gap potentially motivating fewer fine-tuning epochs. In this work, we focus specifically on factors related to the \emph{source} and \emph{target datasets} to address the common imbalance where research efforts prioritize model architectures over data-centric considerations \cite{sambasivan2021everyone,raji2021ai,varoquaux2022machine}.

\begin{table*}[!ht]
    \centering
    \begin{tabular}{p{0.135\textwidth}|p{0.38\textwidth}|p{0.405\textwidth}}
    \toprule
     \textbf{Category} & \textbf{Definition} & \textbf{Example} \\
     \midrule
    \textbf{Source-only: size} & Larger source datasets help learn general features, while sufficient target samples provide effective adaptation. Sample size influences the balance between broad generalization and task-specific learning. & \cite{malik2022youtube}: ``Although not directly related to brain scans, \textbf{the vast array of real-world actions depicted by the images and videos} can provide the basis for a strong, general feature extractor.'' \\ \midrule
    \textbf{Source-only: task complexity} & Refers to the inherent difficulty of a task based solely on the source dataset. It emphasizes how the number and variety of source classes contribute to richer learned representations, which in turn affects transferability. It focuses on balancing representational diversity with task-specific discrimination. & \cite{ribeiro2017exploring}: ``It can be seen in Table 3 that with the \textbf{same number of images and classes, texture databases perform better than natural image databases} specially in the ALOT, CELIAC and DTD databases''. \\ \midrule
    \textbf{Source-target: task complexity similarity} & Refers to the difficulty of transferring knowledge from a source task to a target task, based on the alignment between their data distributions, label semantics, and feature spaces. It captures how well the representations learned from the source domain generalize to the target domain. & \cite{ribeiro2017exploring}: ``in a fair comparison (with the same number of images in all database) \textbf{when the number of classes is the same of the target database (two classes)}, the results are better than using more classes.'' \\ \midrule
    \textbf{Source-target: semantic similarity} & Refers to how closely related the meanings or concepts represented in the source and target datasets are, for example ``human-made objects'' vs ``animals''. The focus is on the underlying meaning rather than visual characteristics. & \cite{chen2019med3d}: ``We believe that the pre-trained model based on \textbf{3D medical dataset should be superior to natural scene video} in 3D medical target tasks'' \\ \midrule
    \textbf{Source-target: visual similarity} & Refers to the extent to which the source and target datasets share perceptual and structural characteristics, such as texture, shape, color distribution, and spatial composition. & \cite{shi2018learning}: ``For the breast imaging tasks, we believe that better representation of deep features can be learned if deep learning models can be trained on more \textbf{similar domains, such as the texture datasets, or medical image datasets on other human body parts}.'' \\ \midrule
    \textbf{Source-target: feature space similarity} & Refers to the degree to which the source and target dataset produce comparable feature embeddings when processed through a shared or pretrained model. It focuses on how aligned the internal representations, such as activation patterns or latent vectors, are across domains. & \cite{yang2023pick}:``we propose a new method using class consistency and feature variety (CC-FV) with an efficient framework to estimate the transferability in medical image segmentation tasks. Class consistency employs the \textbf{distribution of features extracted from foreground voxels of the same category in each sample to model and calculate their distance}, the smaller the distance the better the result;''  \\ 
     \bottomrule
     \end{tabular}
     \caption{Criteria or categories considered by researchers in the adoption of transfer learning. Emphases in the quotes are ours.}
     \label{tab:categories}
\end{table*} 

\subsection{Source-only factors} \label{subsec:source-factors}
It is widely accepted that the \emph{source dataset} size is important to transferability, as both theoretically and empirically we know that more training data leads to better generalization. Of course, this is not simply a question of the number of images - we could replicate the \emph{source dataset} infinitely to increase the ``official'' training size, but there would be no influence on the generalizability of the trained models. The source data therefore needs to be diverse and representative, both currently ill-defined concepts within ML \cite{clemmensen2022data, zhao2024position}. 



The \textbf{task complexity} or learnability of the source classification task is influenced by the \textbf{number of classes} and label granularity. High class overlap from noisy labels or invisible characteristics, such as pneumonia diagnosis which suffers from low annotator agreement \cite{oakden2019exploring}, can result in poor source model performance. Such models are often less useful than those trained on smaller yet more curated datasets. A trade-off exists between sample size and complexity, where general classes like ``cancer'' or ``non-cancer'' provide more examples per class but may be harder to learn if fine-grained categories like melanoma and keratosis have distinct visual characteristics. Consequently, datasets with many fine-grained classes can sometimes outperform general ones even with fewer samples. In similar cases, removing some classes with low sample sizes or high label noise may prevent model confusion.

\subsection{Source-target factors} \label{subsec:source-target-factors}

Considering both \emph{source} and \emph{target datasets}, various other considerations come into play, often related to the ``similarity'' between source and target, which is again an ill-defined concept, as \cite{cheplygina2019cats} shows in a scoping review.


Research considers \textbf{semantic similarity} when both datasets originate from the medical domain \cite{chen2019med3d}, assuming the source model learns features more relevant to the target task. However, selecting a closely related source often involves a trade-off with sample size, as medical datasets are typically much smaller than natural image collections. While early successes in transferring from ImageNet-1K were attributed to its massive scale providing general features beneficial for any classification task, RadImageNet \cite{mei2022radimagenet} was later introduced to bridge this gap. Comprising 1M radiological images, RadImageNet serves as a domain-specific, large-scale alternative that has been shown to outperform ImageNet-1K.



Research also considers \textbf{visual similarity} in terms of perceived textures and structures, regardless of semantic content. For instance, Shi~\etal~\cite{shi2018learning} found that pretraining on ImageNet-1K, the Describable Textures Dataset (DTD) \cite{cimpoi2014describing}, and INBreast \cite{moreira2012inbreast} yielded comparable results, despite DTD and INBreast being significantly smaller than ImageNet-1K. This highlights that visually similar images can be semantically distinct, as exemplified by the chihuahua versus muffin meme, just as semantically different images may share visual characteristics. We visualize these differences between the categories in our context in Fig~\ref{fig:abstract}, where \textit{Source A} and \textit{Target} are semantically similar but visually dissimilar, whereas \textit{Source B} and \textit{Target} are semantically dissimilar but visually similar.


Beyond qualitative perception, similarity can be measured quantitatively through \textbf{feature space similarity}. By embedding datasets into a shared representation space using descriptors like SIFT, HOG, or off-the-shelf extractors, one can quantify distributional closeness via metrics such as Kullback-Leibler divergence to improve alignment. While more explicit before the advent of deep learning, this approach persists implicitly through techniques like intensity normalization. Examples of such measures for both general computer vision and medical imaging are discussed in \cite{juodelyte2024dataset}.

Finally, \textbf{similarity of task complexity} (rather than just complexity of the source task) is sometimes mentioned as a factor contributing to transferability. If the target task has fine-grained labels, researchers have hypothesized that fine-grained source tasks would lead to higher transferability. 

\section{Methods: questionnaire} \label{sec:survey}

\subsection{Questionnaire design} To explore how ML practitioners select \emph{source datasets}, we designed a three-part questionnaire in which Part 1 captures background and experience, Part 2 documents practical choices regarding a recent \emph{source dataset}, and Part 3 conceptualizes tacit knowledge through two controlled case studies. The design was informed by a pilot test with three PhD students and postdoctoral researchers who provided written feedback on survey wording and response options. Table~\ref{tab:questionnaire} provides an overview of the final questionnaire, which is available in full in Appendix \ref{appendix:questionnaire}.

\begin{table*}[ht]
    \centering
    \small
    \renewcommand{\arraystretch}{1.15}
    \begin{tabular}{
    >{\arraybackslash}m{0.03\textwidth} 
    >{\arraybackslash}m{0.13\textwidth} 
    >{\arraybackslash}m{0.74\textwidth}
    }
        \toprule
        \textbf{Part} & \textbf{Main focus} & \textbf{Main items} \\
        \midrule
        \textbf{1} & Background \& experience &
        Position; years of ML experience; primary domain; types of transfer learning; data setting; optional country and email. \\
        \addlinespace
        \textbf{2} & Most recent TL project &
        Project category and main goal; source and target datasets; model design; evaluation methods; reasons for the chosen source; reasons against alternatives. \\
        \addlinespace
        \textbf{3} & Case studies &
        Paired case studies with visually and semantically distinct tasks; several candidate pretraining sources per case; likelihood of choosing, expected performance, matrix ratings on model-level effects, free-text reasons. \\
        \bottomrule
    \end{tabular}
    \caption{Overview of the designed questionnaire.}
    \label{tab:questionnaire}
\end{table*}




The landing page explained that the study examines intuitive pretraining source selection and instructed participants to rely solely on personal experience rather than web searches or AI tools. Email collection remained optional, where participants provided unique case numbers to allow anonymous follow-up.

Part 1 collected participant positions, ML experience, and research domains to describe the sample and control for seniority and domain differences. We documented whether participants used public or private datasets and provided optional fields for country and contact (\emph{i.e.,} emails).

Part 2 documented recent transfer learning projects by identifying categories, goals, datasets, and architectures. Participants reported practical motivations for selecting a \emph{source dataset} from literature-derived examples including visual and semantic similarity, data scale, prior experience, and model availability, with an additional custom field for alternative reasons.


Finally, to probe context-dependent intuitions beyond a single choice for ``medical images'', we presented two controlled case studies. Each study featured a visually and semantically distinct medical imaging task while offering the same candidate \emph{source datasets}. By varying the target task while keeping the source options constant, this design aimed to reveal how researchers' selection criteria and reasoning change depending on the specific context, thus uncovering their underlying heuristics for choosing a \emph{source dataset}.

The case study 1, \texttt{CS-tissue}, is a classification task on colorectal Hematoxylin and Eosin (H\&E) image patches~\cite{ignatov2024nct}, where the objective was to distinguish between different tissue types. The case study 2, \texttt{CS-xray}, is a multi-label classification task on chest X-rays, requiring the identification of common thoracic pathologies~\cite{irvin2019chexpert}. We chose these as the targets datasets because (1) they are among the most widely used medical imaging modalities, (2) they represent clearly distinct visual and semantic categories, and (3) they expose participants to both color and grayscale imaging.


Within each case, participants followed the same sequence of actions: 
\begin{enumerate}
    \item Indicating likelihood of choosing each \emph{source dataset}, with the options \textit{Likely}, \textit{Neutral}, \textit{Unlikely}, and \textit{Not sure};

    \item Assessing the expected fine-tuning performance for each \emph{source dataset} on a five-point Likert scale, where \textit{1} means \textit{Very poor} and \textit{5} means \textit{Very good};

    \item Assessing the expected effects on the resulting model using a matrix that included domain similarity, visual similarity, embedding similarity, dataset scale, fairness, and robustness, and one optional criterion in free text; 

    \item Explaining their choices in a free text field.
\end{enumerate}

\subsection{Datasets and interactive dataset browser}
In the case studies, participants had to judge candidate sources (three potential \emph{source datasets} for a unique \emph{target dataset} per task). 

The \textbf{target datasets} for the case studies \texttt{CS-tissue} and \texttt{CS-xray} were, respectively:
\begin{itemize}
    \item \textbf{CRC-VAL-HE-7K}~\cite{ignatov2024nct}. A collection for 9 class, patch-level tissue classification with 7,180 non-overlapping colorectal H\&E patches from 50 patients with colorectal adenocarcinoma.

    \item \textbf{CheXpert}~\cite{irvin2019chexpert}. This subset contains 834 chest radiographs from 662 unique patients, focusing on eight common thoracic pathologies after classes with fewer than 100 images were removed.
\end{itemize}

For both tasks, participants considered the same three \textbf{source datasets}:
\begin{itemize}
    \item \textbf{ImageNet-1K}~\cite{deng2009imagenet}. A large-scale dataset with 1.3M images of everyday objects and concepts from 1K categories. It serves as a de facto standard for benchmarking computer vision models and pretraining, making it a common baseline in transfer learning research.

    \item \textbf{RadImageNet}~\cite{mei2022radimagenet}. A domain-specific alternative, containing approximately 1.35M radiological images (CT, MRI, Ultrasound) spanning 165 pathologies. Its primary purpose is to improve model performance on medical tasks compared to models pretrained on non-medical data like ImageNet-1K.

    \item \textbf{Ecoset}~\cite{mehrer2021ecologically}. Created to better align object categories with human vision. It contains over 1.5M images of everyday objects and concepts selected based on their relevance to humans and linguistic frequency.
\end{itemize}

To facilitate dataset assessment, we developed an online dataset browser\footnote{\url{https://choice-intuition.streamlit.app/}} for quick visual comparison. We show a screenshot in Figure~\ref{fig-tool} in Appendix \ref{appendix:databrowser}. The tool allowed participants to compare two datasets by displaying the object classes and class sizes within each dataset, and showing a random sample of images from that class. The tool intentionally omitted performance metrics or other metadata to ensure judgments were based solely on visual evidence.

\subsection{Participants and data collection}
We recruited participants through multiple networking platforms and disseminated the information through the research team's professional networks. To reach a larger audience, we also shared the call for participation in Slack channels of specialized communities. Furthermore, we sent direct email invitations to researchers who had previously engaged with our work. Data for this study was collected between August 7th and August 28th, 2025, via a survey hosted on the SoSci Survey platform. Prior to data collection, the study protocol was cleared by the authors' institutional ethics board. 


The study included 15 participants from diverse academic and professional backgrounds. Table~\ref{tab:participants} summarizes their positions and extensive experience in ML. In terms of research backgrounds, the most common area was medical imaging, followed by computer vision, algorithmic fairness, image restoration, and image compression, see Fig.~\ref{fig:wordcloud}. In terms of practical experience, most participants have worked on fine-tuning (93.3\%), feature extraction (73.3\%) and domain adaptation (73.3\%). Regarding datasets, the largest group reported using public datasets (5), followed by an equal use of both public and private datasets (4), and lastly private datasets, i.e., proprietary or internal ones (2). Participants were distributed across the world, including Brazil, China, Denmark, Germany, Israel, The Netherlands, Portugal, Republic of Korea, Spain, Switzerland, and United Kingdom.


\begin{table}[ht]
    \centering
    \begin{tabular}{lclc}
    \toprule
    \textbf{Position} & \textbf{Count} \\
    \midrule
    MSc student & 2 & Associate prof & 1\\
    PhD student & 5 & Full prof & 2\\
    Postdoc & 2 & Research scientist & 1\\
    Assistant prof & 1 & Industry & 1\\
    \bottomrule
    \end{tabular}
    \caption{Participant statistics. Participants reported a mean of 8.9 years of ML experience (median 7.0, IQR 3.5--15.0) and 6.5 ML papers (median 3.0, IQR 1.0--8.5).}
    \label{tab:participants}
\end{table}

\subsection{Quantitative analysis}
We organized the collected data by participants' unique Case ID, where we removed one participant who entered an impossible number for the ``years of experience'' and entered the same word for all open questions. 



We used stacked Likert charts to visualize the distributions of willingness and fine-tuning performance for each case and dataset. The charts showed the percentage at every response level, including \textit{Not sure}. This respects the ordinal scale, avoids assumptions about means, and makes differences across datasets and cases easy to observe.

Please note that we are aware of the small sample size in the survey. However in the light of replicability for future studies, we still report the types of statistical significance tests used for assessing the expected performance and multidimensional assessment questions. We do not base our conclusions on the p-values of the tests.

For expected performance, we treated respondents as paired and used the Friedman test to assess overall differences across datasets. If the overall test indicated differences, we ran pairwise Wilcoxon signed-rank tests with Holm correction to control for multiple comparisons. We reported effect sizes using Kendall’s $W$ for the overall comparison and $r$ for each pairwise contrast. This matches a repeated-measures setting with ordinal data and a small sample size while keeping the results easy to interpret. For multidimensional assessment, we computed Spearman rank correlations between each dimension and expected performance. We reported the correlation coefficient $\rho$. This test fits ordinal or skewed data and is robust to outliers. It shows which dimensions move together with expected performance and which move in the opposite direction.

\subsection{Qualitative analysis}
In analysis of the qualitative answers, we followed the Directed Content Analysis \cite{hsieh_three_2005}. This approach enabled analyzing qualitative responses using theoretical insights from prior work, while remaining open to new factors capturing candidates' intuition.

Our literature review on factors influencing transfer learning (Section \ref{sec:factors}) served as the entry point to coding. Based on these factors, <two anonymized authors> jointly developed a codebook (see Table~\ref{tab:codebook}). Each code (N=15) was described through its definition, guidance on when to apply or not apply it, and an example \cite{thompson_guide_2022}. The initial set covered literature-based factors while leaving room for emergent codes. The same authors then independently coded all open-ended responses to the case studies (Q19 and Q23), applying the predefined codes and introducing new ones where necessary. They subsequently met to compare their usage of codes, resolve discrepancies, and refine the inductive codes. The data were then revisited with the updated codebook for consistency, followed by a final discussion to align the coding across authors and responses. Once finalized, we quantified code frequencies across responses and examined how these patterns related to the quantitative results, enabling a richer, mixed-methods interpretation.

\section{Results} \label{sec:results}

\subsection{Quantitative results}
\textbf{Project type.} Projects were mainly concentrated in medical imaging (40.0\%) and image classification (33.3\%), followed by other types (20.0\%). Semantic segmentation accounted for 6.7\%. No responses were recorded for the remaining predefined categories.

\noindent \textbf{Goal of the project.} The most common aims were to improve performance on the target task (60.0\%), improve robustness or generalization (46.7\%), and adapt to a new domain (40.0\%). Reducing training time or data was selected by 26.7\%. Smaller shares reported exploring the feasibility of transfer learning or other goals (13.3\% each).



\begin{figure*}[ht]
    \centering
    \includegraphics[width=\textwidth]{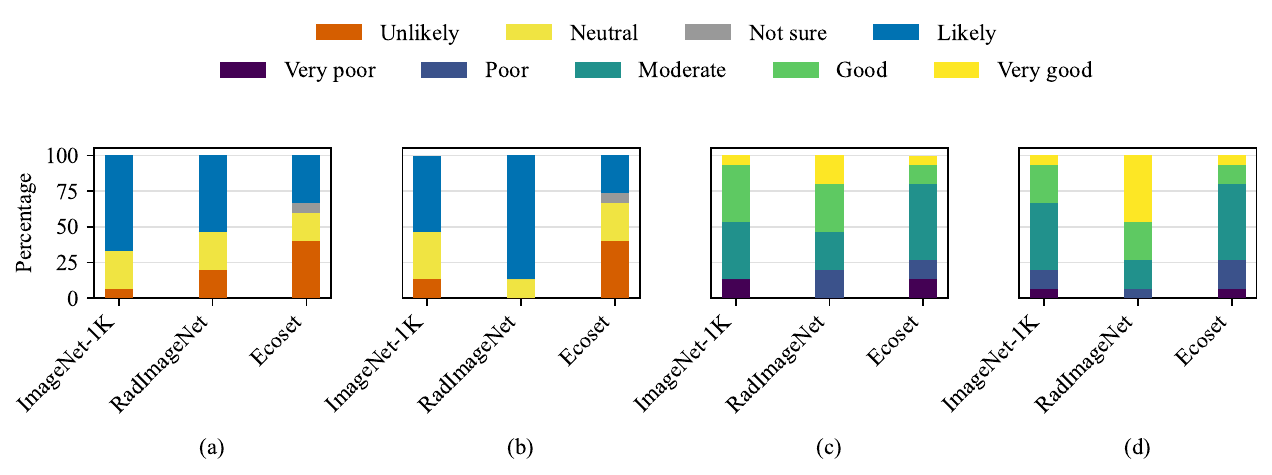}
    \caption{Participants' willingness to choose each source dataset and subjective expectations of fine-tuning performance per source. (a)(c) \texttt{CS-tissue}, (b)(d) \texttt{CS-xray}.}
    \label{fig:willingness}
\end{figure*}

\noindent \textbf{Willingness to choose a dataset source.} Overall, practitioners prioritized practical factors, such as dataset scale and pretrained model availability, over visual or semantic similarity. Experience-based reasons were less common. Fig.~\ref{fig:willingness} shows participants' willingness to use each source for the two case studies. For \texttt{CS-tissue}, ImageNet-1K remained the most preferred source. For \texttt{CS-xray}, however, preferences shifted distinctly toward the medical source (RadImageNet). Ecoset consistently ranked last in both cases.

\noindent \textbf{Expected fine-tuning performance.} The choices for the expected fine-tuning performance are shown in Fig.~\ref{fig:willingness}. Overall expectations were broadly positive across both cases. However, a clear domain-specific shift emerged. In \texttt{CS-tissue}, expectations were relatively similar across sources, reflecting the strong influence of model availability and common use. By contrast, \texttt{CS-xray} shows a strong preference for the medical source in a clearly defined medical task, where expectations for RadImageNet shifted noticeably upward compared to \texttt{CS-tissue}. These aspects may help RadImageNet and ImageNet-1K, while Ecoset receives fewer high expectations, see Fig.~\ref{fig:willingness}. Expectations for ImageNet-1K and Ecoset remained relatively stable across the two cases.

\noindent \textbf{Other expected effects beyond performance.} 

We relate expected fine-tuning performance to the ratings on six dimensions (dataset scale, embedding similarity, visual similarity, domain similarity, fairness, and robustness) to observe their descriptive trends (see Fig.~\ref{fig:radar} in the Appendix). In \texttt{CS-tissue}, performance expectations were most strongly associated with embedding and domain similarity. In \texttt{CS-xray}, domain alignment became the primary driver of expectations, particularly for ImageNet-1K and Ecoset, while RadImageNet consistently yielded high performance expectations regardless of variance in perceived similarity. Notably, across both cases and all sources, dataset scale and fairness evaluations did not appear to systematically drive performance expectations. This yields the expected ordering indicating respondents expect the medical source to fine-tune best for chest X-ray tasks, see Fig.~\ref{fig:willingness}.
\subsection{Qualitative results}
Based on our qualitative analysis, we identified three overarching categories that influence researchers' choices when selecting \emph{source datasets} for the two case studies:

\textbf{Research community influence.}
Researchers often rely on personal experience (``based on my experience''), peer recommendations (``I heard from colleagues''), and established community practices (``widely adopted''). External incentives also play a role, such as reviewer expectations (``must be tested as a baseline'', ``reviewers might ask''). 

\textbf{Attributes of the source dataset.}
Practical considerations such as ease of use and prior familiarity influence selection (``easy to use'', ``I never worked with this dataset, so I would not select it''). The availability of pretrained models and the datasets' popularity also matter. A few participants were unaware of lesser-known datasets like Ecoset. Participants highlighted size and diversity as two source qualities (``large-scale, diverse visual data''), which are linked to the ability to learn transferable features  (``models to learn transferable low- and mid-level features'').

\textbf{Similarity between source and target datasets.}
This includes both semantic and visual similarity. Some participants expressed skepticism about domain mismatch (``natural image dataset would not perform well on chest X-rays'') and support for domain alignment (``large-scale medical dataset may provide more relevant features''). Some participants also expressed skepticism about using datasets from the same domain (medical imaging) but of a different modality (``More similar as medical images, but different modality from histology.'').  Visual similarity was discussed in terms of color (``The images are RGB'', ``Colour images are easier to transfer'') and structural features like texture and shape (``large part of the image is background'', ``models learn to recognize edges, shapes'').

Additionally, we identified two residual categories: (1) Unspecified \emph{source dataset} qualities. Participants referred to attributes like ``good image quality'' without further elaboration. (2) Unspecified domain similarity: Terms like ``domain gap'' or ``more similar'' were used without clear definitions.

\textbf{The uncomfortable absentee: fairness}. 
Fairness and bias considerations were nearly absent from participants' reasoning about source dataset selection. Across all collected reasons for dataset rejections, they appeared only twice: once questioning model reliability (``my experience is that this kind of models are quite OK since they learn useful features. However, they may not be very reliable''), once noting a potential bias toward specific object categories (``seems to have a bias towards specific object categories''). More tellingly, one participant signaled readiness to compromise on fairness, in the name of expected performance gains: ``recent foundational models trained in TCGA have outperformed the rest of the models. Therefore, it will be my `go-to' dataset. However, I have some concerns about fairness and generalization.'' This suggests that while fairness and bias are acknowledged, they are secondary to other concerns, such as performance or community influence. Given that source dataset choices propagate latent biases forward into deployed models, practitioners are effectively making consequential fairness decisions by omission, without necessarily recognizing them as such.

\subsection{Alignment of quantitative and qualitative results} In \texttt{CS-tissue}, dimensional patterns aligned with qualitative explanations, as expected performance correlates most with embedding similarity, followed by semantic and visual similarity. These correlations suggest a coherent pattern, though they may reflect consistency in self-reported judgments. In \texttt{CS-xray}, quantitative findings aligned with qualitative accounts where RadImageNet was expected to outperform alternatives due to domain alignment, pretrained model availability, and established community baselines.

Some mismatches persisted because similarity ratings and expected performance for RadImageNet did not always move together, suggesting that imaging modality differences and heterogeneous content weaken the ``more similar is better'' relationship. Lower expectations for Ecoset aligned with reported unfamiliarity. Dataset size was noted but rarely influenced expectations unless prioritized by specific project goals. Finally, fairness considerations neither shaped performance expectations nor drove source dataset selection, suggesting a secondary role in practitioners' decision-making. Overall, expectations were shaped primarily by perceived domain fit, model availability, and community practices.

\section{Discussion}

While researchers’ intuitions about ML transferability may be loosely inspired by human cognitive experiences, such analogies can be misleading given the fundamental differences between neural networks and the human brain. Rather than relying on these informal cognitive parallels, this study provides a systematic investigation into the specific factors that guide practitioner decision-making.

\subsection{From experiments to intuitive insights} Our categorization of transfer learning factors builds upon prior studies in computer vision and medical imaging. These studies typically explore source-only or source-target factors such as dataset size, number of classes, model complexity, and fine-tuning strategies \cite{shin2016deep, raghu2019transfusion, minaee2020deep}. Some have examined semantic differences, including the impact of pretraining on general vs. domain-specific (medical imaging) datasets \cite{malik2022youtube,chaves2023performance,shi2018learning}. Unlike controlled experiments that systematically vary individual factors, our study elicited practitioners' intuitions across realistic decision scenarios. However, this realism imposed its own constraints. By keeping the \emph{source dataset} scale roughly constant (between 1M and 1.3M images), we minimized the effect of dataset size as a distinguishing factor. This design choice allowed us to focus on other (potentially unknown) dimensions of decision-making when selecting a dataset for transfer learning. This choice can also explain the lack of correlation between the dataset scale and expected performance among our participants, which may be a result of the case study setup rather than participants genuinely discarding this property.

\textbf{Practitioners' intuitions are contextual and hierarchical.} Our two target tasks created distinct conditions for evaluating similarity. \texttt{CS-tissue} emphasized embedding, visual, and domain similarities ($\rho=0.9$, $\rho=0.8$, and $\rho=0.7$). On the other hand, \texttt{CS-xray} suggested, in line with previous research, that domain alignment primarily drives dataset selection ($\rho=0.7$). This trend also appeared in the distribution of expected likelihood of performance improvement: RadImageNet $86.7\%$, ImageNet-1K $53.3\%$, and Ecoset $26.7\%$. This variation confirms that practitioners' intuitions are contextual and hierarchical. When a clear domain match is available (radiological images for chest X-rays), domain similarity becomes the primary decision criterion, overshadowing other factors. However, when domain alignment is ambiguous or absent (radiological images for histopathology), practitioners draw on alternative heuristics to differentiate between source options. Further investigation is necessary to determine whether a wider range of medical imaging tasks (e.g., ultrasound, dermoscopy, retinal imaging) would reveal distinct patterns of reasoning.

\textbf{Tacit knowledge anchors transfer learning decisions.} Most importantly, our focus on self-reported similarity perception and performance expectations, rather than computer transferability metrics or actual transfer learning results, captures the intuitive mental models guiding real-world transfer learning decisions. These tacit models act as \emph{heuristic filters} for navigating a large and complex decision space. This means that the misalignments, e.g., ImageNet-1K having the highest likelihood of use in \texttt{CS-tissue}, but not the best expected performance outcome, reflect the multidimensionality of practitioners' decision-making. Our survey brings complementary knowledge to existing efforts aimed at understanding transfer learning from the perspective of ML researchers. Recent work has explored the conceptualization of the tacit knowledge of data practitioners, such as integrating human intentions into fine-tuning workflows \cite{zeng_intenttuner_2024} or examining how non-expert users engage with transfer learning \cite{mishra_designing_2021}. Our analysis identified novel factors influencing researchers' decision-making in the context of transfer learning in medical imaging. These include personal experiences, recommendations from colleagues, established community practices, and external incentives such as reviewer expectations.

Earlier work on dataset choice \cite{cheplygina2019cats} identifies two main lines of reasoning on which choices lead to robustness: large, general datasets or domain-specific ones. Our survey follows the same trend, it is not entirely clear why some researchers prefer one or the other. ``Previous experience'' is noted as an important factor, but this could encompass many different types of experience, from empirical evidence in the researcher's own project through discussions with colleagues to findings in the literature. In all cases, we suspect that there are biases at play, such as ``historical bias'' on what type of data the research group studies, publication bias in favor of successful results, and so on. This is not to say that there is a single ``ground truth'' answer to which option is better overall (also known as ``no free lunch'' in ML), and this is going to be dependent on the domain context of the target task. Yet it should be feasible to derive rules of thumb for which options might be more promising, giving a starting point for researchers addressing previously unexplored tasks. Our recommendation for ML researchers is therefore to document their decisions (what they do and why) to support responsible research. Clear reporting of choices (e.g., \emph{source dataset}) can enhance accountability and ethical awareness, helping the community understand potential biases and limitations in the work \cite{gebru2021datasheets, mitchell2019model}.

\textbf{Reasoning is shaped by experience but vulnerable to bias.} Our focus on self-reported similarity perception and performance expectations captures the intuitive mental models guiding real-world transfer learning decisions. These tacit models act as \emph{heuristic filters} for navigating complex decision spaces, which explains misalignment such as ImageNet-1K having the highest likelihood of use in \texttt{CS-tissue} despite suboptimal performance expectations. While these findings complement research on the tacit knowledge of data practitioners \cite{zeng_intenttuner_2024, mishra_designing_2021}, they also reveal a troubling gap where even experienced practitioners lack the conceptual vocabulary and decision frameworks needed for principled source selection. Our analysis identified novel factors like personal experience and reviewer expectations, but the absence of feature-space similarity reasoning despite its prominence in literature \cite{nguyen2020leep, juodelyte2024dataset} demonstrates that methodological advances remain siloed from deployment decisions. Consequently, developing better metrics or larger datasets is insufficient without parallel efforts to translate these advances into accessible heuristics and tools that foreground equity considerations.


\subsection{Challenges in operationalizing metrics in transfer learning}
It is essential to emphasize that studying the broader implications of data in ML, rather than merely inventing new methods, is vital to ensuring high-performing and responsible ML development. The choices made in research, from problem framing to dataset selection, are never neutral. They encode specific values that shape societal outcomes \cite{birhane2021values}. Moreover, \JimSan~\etal~\cite{jimenez2025picture} showed how datasets are not static but ``living'' entities that condition performance and generalizability of ML models. As noted by Sambasivan~\etal{}~\cite{sambasivan2021everyone}, the undervaluation of data work perpetuates systemic biases and overlooks the labor and context necessary for meaningful ML systems.

\textbf{Community norms influence source dataset selection.} We build on this perspective by revealing how \emph{source dataset} selection, a critical but understudied step in transfer learning, is shaped by community norms rather than solely technical and academic standards. We found that ImageNet-1K maintains a strong preference (66.7\% and 53.3\% likely across our cases) despite participants recognizing domain misalignment and not expecting the most performance gains. Our participants ascribed this preference to reviewer expectations and baseline requirements: ``reviewers might ask,'' ``must be tested as a baseline.'' These soft-power expectations can lead to suboptimal choices. For example, even though RadImageNet matched the chest X-ray domain (86.7\% likely, $\rho=0.7$ for domain similarity), practitioners still felt compelled to compare against ImageNet-1K, potentially wasting computational resources on a known, inferior baseline rather than exploring alternative medical sources.

\textbf{Implicit similarity definitions may affect equity.} In free-text responses, several concepts emerged without sufficient context or precise definitions, particularly those related to quality and similarity, such as ``domain mismatch'' and ``domain gap''. These terms were often used in ambiguous comparisons like ``more/less similar,'' without specifying whether the similarity referred to visual features (e.g., color, texture, shape) or semantic content. This conceptual ambiguity echoes concerns raised in prior work, Clemmensen~\etal{}~\cite{clemmensen2022data} proposed a coding framework to systematically capture notions of representativity, while Zhao~\etal{}~\cite{zhao2024position} provided guidance for defining and evaluating dataset \textit{diversity}, both emphasizing the need for precise conceptual definitions when reasoning about datasets. Without explicit definitions, implicit assumptions may reflect local institutional demographics, risking building models that potentially fail underrepresented populations~\cite{seyyed-kalantari_underdiagnosis_2021}. 

These findings resist the critique that ML practitioners focus solely on technical optimization. Our results show that social factors shape source dataset selection: community norms and institutional expectations carry real weight in practitioners' reasoning. Fairness, however, is structurally different. A model that underperforms or departs from community standards is penalized immediately and visibly~\cite{hutchinson2022evaluation}. A model that encodes demographic bias is not, at least not at the point of dataset selection, and rarely through mechanisms that practitioners personally experience. Rather, it has downstream patient-safety consequences~\cite{gichoya_equity_2021}. This asymmetry in consequences explains why fairness surfaces as an acknowledged but deprioritized concern: practitioners are not indifferent to it, but they lack the incentive structures and evaluation frameworks that would make it actionable~\cite{madaio2020co, holstein2019improving}. Treating fairness as an optional consideration, one that can be incorporated when convenient and set aside when it conflicts with performance or community standards, is insufficient~\cite{kim2020fact, rodolfa2021empirical}. For fairness to matter in transfer learning practice, it must become a structural requirement, assessed with the same specificity and accountability as model performance.

\subsection{Limitations} While we focused on a crucial subset of factors influencing transfer learning, we did not quantify each factor's contribution through experiments and instead suggest experimental validation for future research. 
Although limited by sample size, our quantitative analysis focuses on reporting trends in the findings, and it remains appropriate for a qualitative study in this under-explored space, particularly given the niche population. Also, the large proportion of academic participants makes this study more valuable for foundational research than for developing production-ready systems. Furthermore, potential anchoring bias may have resulted from the fixed question order across case studies, as earlier responses could have shifted judgments on subsequent items.

Our findings are specific to medical imaging classification and may not generalize to other ML domains because medical imaging is not simply ``small computer vision \cite{jimenez2024copycats}''. Evidence shows that advances in general computer vision often fail to translate directly to medical applications \cite{raghu2019transfusion,mei2022radimagenet,juodelyte2023revisiting}, underscoring the continued need for domain-specific investigation alongside related fields. Consequently, our results are restricted by the modalities covered and the specific nature of the classification tasks examined.

\subsection{Concluding remarks} As ML models in data-scarce domains increasingly rely on transfer learning, understanding how researchers choose \emph{source datasets} is essential for shaping successful outcomes. We introduced a conceptual framework (Section~\ref{sec:factors}) and conducted a task-based survey (Section~\ref{sec:survey}) to surface the tacit knowledge and heuristics guiding these selection processes. Our findings (Section~\ref{sec:results}) reveal that practitioners rely on intuition, personal experience, and community norms such as reviewer expectations and established baselines even when they acknowledge that such intuitions may be unreliable. By comparing qualitative and quantitative data, we identified limitations in the ``more similar is better'' approach and highlighted how dataset selection is driven by technical performance and social dynamics, while ethical and fairness considerations remain secondary due to a lack of actionable incentives. Furthermore, the frequent yet vague use of concepts like ``domain gap'' and ``good image quality'' underscores the necessity for HCI-focused tools to operationalize these notions. These insights support the development of more deliberate and reflective dataset practices through clearer frameworks and explicit terminology.

\section*{Acknowledgments}
This project has received funding from the Independent Research Council Denmark (DFF) Inge Lehmann 1134-00017B, and from the Novo Nordisk Foundation NNF21OC0068816.

\section*{Statements}
\indent\textbf{Generative AI usage statement}. The following Generative AI tools were used solely for copy-editing and language refinement: Copilot, Claude, ChatGPT, and Gemini. We confirm that the content of this manuscript is our own. 

\textbf{Ethical consideration statement}.
This study received prior approval from the lead author's university’s institutional ethics review board, and all data collection was conducted in accordance with the approved protocols.

\textbf{Researcher positionality statement}.
We approach this research as scholars trained in computer science, human-computer interaction and medical image analysis, but with experience working in interdisciplinary teams with clinicians, lawyers and policymakers. While based at universities in Western Europe, our team brings diverse personal backgrounds to this work. The design of this study stems from a belief in systematic understanding over anecdotal intuition in transfer learning for medical image classification. This perspective guided our focus on an under-explored sociotechnical aspect of ML development, reflecting our dedication to accountability, transparency and reproducibility.

\textbf{Adverse impact statement}.
The goal of our work is to support machine learning researchers in transfer learning by improving transparency and accountability in decision-making for medical image classification. However, the limited sample size restricts our ability to draw definitive quantitative conclusions; therefore, our findings should be interpreted as indicative trends rather than conclusive evidence. 
We investigated two case studies in medical imaging, and further validation across additional applications and modalities is necessary. While our insights and recommendations build on foundational transfer learning research, they are not intended for direct use in clinical decision-making.

\bibliography{refs}

\clearpage 
\onecolumn 
\appendix 

\counterwithin{figure}{section}
\counterwithin{table}{section}
\renewcommand\thefigure{\thesection\arabic{figure}}
\renewcommand\thetable{\thesection\arabic{table}}

\section{Transfer learning notions: paper annotations} \label{appendix:factors}
We show examples (see Table~\ref{tab:annotations}) of prior literature in ML in medical imaging, that discusses different characteristics influencing the transfer learning performance. We used these works as inspiration for defining our initial dimensions, which we then used for our questionnaire. Please note that in this initial search, we only considered these factors as ``present'' (indicated by a check mark) or ``absent'', while in annotations of the questionnaire answers, we distinguished between ``positive'' and ``negative'' effect when the factor was ``present''. The bold emphases in the quotes from the papers are ours. 

\newcommand*\circled[1]{\tikz[baseline=(char.base)]{
    \node[shape=circle, draw, inner sep=0.25pt, minimum size=1em] (char) {#1};}}

\begin{longtable}{p{0.68\linewidth}|c|c|c|c|c|c}
    \toprule
    \multirow{2}{*}{Quote} & \multicolumn{6}{c}{Categories} \\ \cmidrule{2-7}
     & \rotatebox[origin=r]{90}{Semantic similarity} & \rotatebox[origin=r]{90}{Visual similarity} & \rotatebox[origin=r]{90}{Sample size} & \rotatebox[origin=r]{90}{Number of classes} & \rotatebox[origin=r]{90}{Task complexity} & \rotatebox[origin=r]{90}{Model complexity} \\
    \midrule
    \endfirsthead

    \caption*{Examples of considerations influencing transfer learning performance (continued)} \\
    \toprule
    \multirow{2}{*}{Quote} & \multicolumn{6}{c}{Categories} \\ \cmidrule{2-7}
     & \rotatebox[origin=r]{90}{Semantic similarity} & \rotatebox[origin=r]{90}{Visual similarity} & \rotatebox[origin=r]{90}{Sample size} & \rotatebox[origin=r]{90}{Number of classes} & \rotatebox[origin=r]{90}{Task complexity} & \rotatebox[origin=r]{90}{Model complexity} \\
    \midrule
    \endhead

    \midrule
    \multicolumn{7}{r}{\textit{continued on next page}} \\
    \endfoot

    \bottomrule
    \noalign{\vspace{10pt}}
    \caption{Examples of considerations influencing transfer learning performance in previous medical imaging literature, which served as the initial formulation of our conceptualization of factors in Section \ref{sec:factors}.}
    \label{tab:annotations}
    \endlastfoot

    Jain~\etal~\cite{jain2023data}: \circled{1} ``As one might expect, not all source classes have large influences. Figure 1 displays the most influential classes of ImageNet with CIFAR-10 as the target task. Notably, \textbf{the most positively influential source classes turn out to be directly related to classes in the target task} (e.g., the ImageNet label “tailed frog” is an instance of the CIFAR class “frog”). ... Interestingly, the source dataset also contains classes that are overall negatively influential for the target task (e.g., “bookshop” and “jigsaw puzzle” classes).'' & \checkmark &&&&& \\
    \midrule
    Chen~\etal~\cite{chen2019med3d}: \circled{1} ``we believe that the pre-trained model based on \textbf{3D medical dataset should be superior to natural scene video} in 3D medical target tasks.'' & \checkmark &&&&&\\ 
    \midrule
    Tajbakhsh~\etal~\cite{tajbakhsh2016convolutional}: \circled{1} ``we observed a marked performance gain using deeply fine-tuned CNNs, particularly for polyp detection and intima-media boundary segmentation, probably because of the substantial difference between these applications and the database with which the pre-trained CNN was constructed. However, we did not observe a similarly profound performance gain for colonoscopy frame classification, which we attribute to the \textbf{relative similarity between ImageNet and the colonoscopy frames} in our database.'' & \checkmark &&&&& \\
    \midrule
    Menegola~\etal~\cite{menegola2017knowledge}: \circled{1} ``We expected that transfer learning from a \textbf{related task (in our case, from Retinopathy, another medical classification task) would lead to better results}, especially in the double transfer scheme, that had access to all information from ImageNet as well. \textbf{The results showed the opposite}, suggesting that adaptation from very specific — even if related — tasks poses specific challenges.'' & \checkmark &&&&& \\ 
    \circled{2} ``The results suggest that the experimental design is sensitive to the choice of lesions to compose the positive and negative classes, maybe due to the relative \textbf{difficulty of identifying each of the types of cancer evaluated} (Melanomas and Carcinomas).'' &&&&& \checkmark & \\
    \midrule
    
    Cherti~\etal~\cite{cherti2022effect}: \circled{1} ``we conduct a series of large-scale pre-training and transfer experiments where \textbf{we vary not only ResNet model and dataset size during pre-training, but also the domain of the source and the target datasets}, being either natural or medical X-Ray chest images, which allows us to study effect of scale on both intra- and inter-domain transfer.'' & \checkmark & & & \checkmark & & \\
    \midrule
    Raghu~\etal~\cite{raghu2019transfusion}: \circled{1} ``A performance evaluation on two large scale medical imaging tasks shows that surprisingly, transfer offers little benefit to performance, and simple, \textbf{lightweight models can perform comparably to ImageNet architectures}.'' & \checkmark  & & & && \checkmark \\
    \circled{2} ``The results, in Table 3, suggest that while transfer learning has a \textbf{bigger effect with very small amounts of data}, \textbf{there is a confounding effect of model size} – transfer primarily helps the large models (which are designed to be trained with a million examples) and smaller models again show little difference between transfer and random initialization.'' &&& \checkmark &&& \checkmark \\
    \midrule
    Lei~\etal~\cite{lei2018deeply}: \circled{1} ``we utilize a cross-model transfer learning strategy since the two datasets (i.e., ICPR2012 and ICPR2016-Task 1) \textbf{not only are similar in terms of the low-level features, but also are alike in the high-level classification features}.'' & \checkmark & \checkmark &&&& \\
    \midrule
    Xie~\etal~\cite{xie2018pre}: \circled{1} ``We hypothesize that the network pre-trained on grayscale images has the potential to learn more \textbf{features relevant to grayscale images}, which serves to boost the transfer learning performance when applied to a grayscale medical dataset.'' & \checkmark & \checkmark &&&& \\
    \midrule
    Shi~\etal~\cite{shi2018learning}: \circled{1} ``For the breast imaging tasks, we believe that better representation of deep features can be learned if deep learning models can be trained on more \textbf{similar domains, such as the texture datasets, or medical image datasets on other human body parts}.'' & \checkmark & \checkmark &&&& \\
    \circled{2} ``we observed that our best classification performance is from deep features extracted at the middle level layer, ..., deep features at middle-level layers are also regarded to be associated with different textural patterns. This agrees with the findings from our previous study that \textbf{texture-related computer vision features} were among the most frequently selected for this task.'' && \checkmark &&&& \\
    \midrule
    
    Mensink~\etal~\cite{mensink2021factors}: \circled{1} ``Transfer learning is omnipresent in computer vision. ... Intuitively, the reason for this success is that the network learns a \textbf{strong generic visual representation}, providing a better starting point for learning a new task than training from scratch.'' && \checkmark &&&& \\ 
    \circled{2} ``When \textbf{a target dataset is very large, the effect of transfer learning is likely to be minimal}: all the required visual knowledge can be gathered directly from this target dataset. ... ``A source model trained on a larger dataset is likely to be more beneficial for transfer learning. '' &&& \checkmark &&& \\
    \midrule
    Geirhos~\etal~\cite{geirhos2018imagenet}: \circled{1} ``This is in line with the intuition that for object detection, \textbf{a shape-based representation is more beneficial than a texture-based representation}, since the ground truth rectangles encompassing an object are by design aligned with global object shape.'' && \checkmark &&&& \\
    \midrule
    Ribeiro~\etal~\cite{ribeiro2017exploring}: \circled{1} ``On the basis of the good results obtained compared to the classical features we can conclude that the CNN’s have a good generalization capability for the transfer learning specially using \textbf{texture databases} and with the fine-tuning approach.'' && \checkmark &&&& \\
    \circled{2} ``We also showed that when the texture database for the CNN trained is also limited, the fine tuning with a \textbf{bigger database} can be a good alternative to surpass this problem even with a completely different original database since the number of images is very high.'' && \checkmark & \checkmark &&& \\
    \circled{3} ``It can be seen in Table 3 that with the \textbf{same number of images and classes, texture databases perform better than natural image databases} specially in the ALOT, CELIAC and DTD databases.'' && \checkmark & \checkmark & \checkmark & & \\
    \circled{4} ``It also can be noted that, in a fair comparison (with the same number of images in all database) \textbf{when the number of classes is the same of the target database (two classes)}, the results are better than using more classes.'' &&& \checkmark & \checkmark && \\
    \midrule
    
    Wong~\etal~\cite{wong2018building}: \circled{1} ``In our framework, instead of a classification task which involves complex and abstract concepts such as disease categories, we first train the machine to perform a segmentation task which involves \textbf{simpler concepts such as shapes and structures}'' && \checkmark &&&& \\
    \circled{2} ``There are several limitations of using ImageNet pre-trained CNNs on medical image analysis... the size of the pretrained model may be unnecessarily large for medical image applications. Using \textbf{VGGNet} as an example, its architecture was proposed to \textbf{classify 1000 classes of non-medical images}. Such a large number of classes is \textbf{uncommon in medical image analysis} and thus such a large model may be unnecessary.'' &&&& \checkmark && \checkmark \\
    \circled{3} ``By using a \textbf{segmentation network pre-trained on similar data as the classification task}, the machine can first learn the simpler \textbf{shape and structural} concepts before tackling the actual classification problem which usually involves more complicated concepts.'' && \checkmark &&&\checkmark & \\
    \circled{4} ``There are several limitations of using ImageNet pre-trained CNNs on medical image analysis... \textbf{the size of the pretrained model may be unnecessarily large for medical image applications}. Using \textbf{VGGNet} as an example, its architecture was proposed to classify 1000 classes of non-medical images. Such a large number of classes is uncommon in medical image analysis and thus such a large model may be unnecessary.'' &&&&&& \checkmark \\
    \midrule
    Minaee~\etal~\cite{minaee2020deep}: \circled{1} ``Transfer learning is mainly useful for tasks where enough \textbf{training samples are not available to train a model from scratch}, such as \textbf{medical image classification for rare or emerging diseases}. ... To overcome the limited data sizes, transfer learning was used to fine-tune four popular pre-trained deep neural networks on the training images of COVID-Xray-5k dataset.'' && \checkmark & \checkmark &&& \\
    \midrule
    Malik~\etal~\cite{malik2022youtube}: \circled{1} ``Although not directly related to brain scans, \textbf{the vast array of real-world actions depicted by the images and videos} can provide the basis for a strong, \textbf{general feature extractor}. By applying transfer learning in combination with the largest biomedical dataset in the world in the UKBB, we show improved DNN predictions out-of-sample.'' && \checkmark & \checkmark &&& \\ 
    \circled{2} ``\textbf{The data scarcity} in brain-imaging presents a major challenge to effectively train DNNs in many mission-critical settings. We used emerging transfer learning techniques that learned structured a-priori knowledge (inductive biases) from general purpose datasets: the massive video databases Youtube and the natural images from reference dataset ImageNet.'' &&& \checkmark &&& \\
    \midrule
    
    Chaves~\etal~\cite{chaves2023performance}: \circled{1} ``Label-based methods shows superior results in out-of-distribution scenarios. Out-of-distribution scores might be inflated for binary tasks due to the \textbf{distribution concentration on a single class, and the low number of classes benefits in favor of high transferability scores}. Such an issue is absent in the available benchmarks because the general-purpose classification datasets present many classes and consider transferring from ImageNet as standard practice.'' &&&& \checkmark && \\
    \midrule
    Li~\etal~\cite{li2025well}: \circled{1} ``We find that the \textbf{pretext task of segmentation itself can enhance the model capability of segmenting novel classes}. The benefit of same-task transfer learning, i.e., segmentation as pretext and target tasks, is much more straightforward and understandable than other pretext tasks such as contextual prediction, mask image modeling, and instance discrimination.'' &&&&& \checkmark & \\
    \midrule
    Chen~\etal~\cite{chen2019med3d}: \circled{1} ``Together with the evidence shown in Figure 6 that the training losses of different networks are reduced to a similar level after long-enough training epochs, we can conclude that the extracted features from Med3D networks are \textbf{better generalized for the classification task with a small set of data}, while the other two methods show overfitting issues.'' &&& \checkmark && \checkmark & \\ 
    \circled{2} ``This demonstrates the effectiveness of the learned features of Med3D, which are \textbf{also helpful for the classification task}. Moreover, when the \textbf{network depth is gradually increased, the performance of Med3D also increases}.'' &&&&&& \checkmark \\
    \midrule
    Shin~\etal~\cite{shin2016deep}: \circled{1} ``we explore and evaluate \textbf{different CNN architectures varying in width} (ranging from 5 thousand to 160 million parameters) \textbf{and depth} (various numbers of layers), ... and discuss when and why transfer learning from pre-trained ImageNet CNN models can be valuable'' &&&&&& \checkmark \\
    \midrule
    Ke~\etal~\cite{ke2021chextransfer}: \circled{1} ``we find that, for models without pretraining, \textbf{the choice of model family influences performance more than size} within a family for medical imaging tasks.'' ... ``we observe that ImageNet pretraining yields a statistically significant boost in performance across architectures, with a \textbf{higher boost for smaller architectures}.'' &&&&&& \checkmark \\
\end{longtable}

\section{Interactive dataset browser} \label{appendix:databrowser}
\begin{figure}[!ht]
    \centering
    \includegraphics[width=0.7\linewidth]{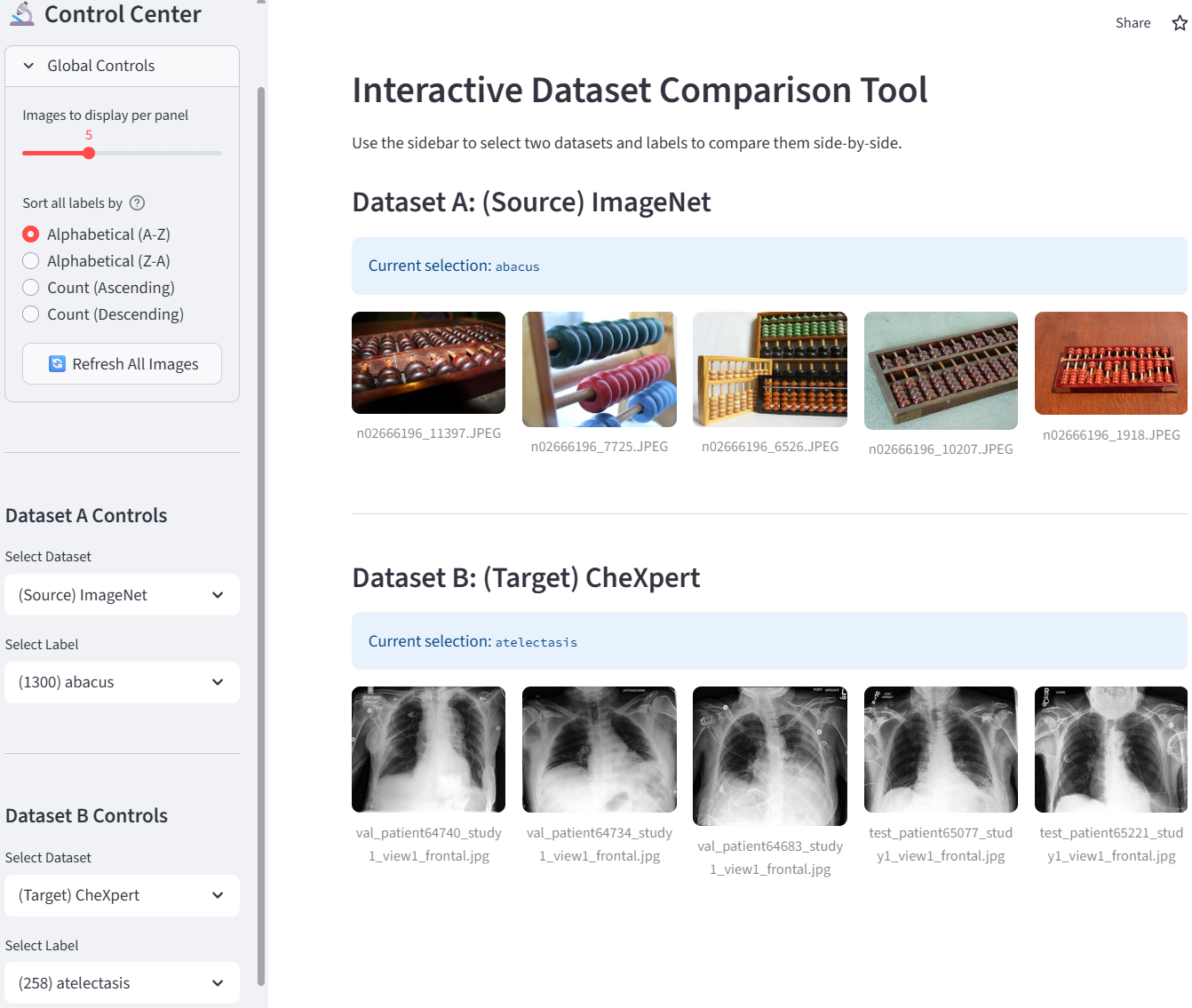}
    \caption{Screenshot of our interactive dataset browser. Users can select a dataset and a label from panels A and B to view random thumbnails. The control panel allow the user to sort labels and refresh the images.}
    \label{fig-tool}
\end{figure}

To aid participants in assessing possibly unfamiliar datasets, we developed an online dataset browser (\url{https://choice-intuition.streamlit.app/}) for quick visual comparison. As illustrated in Figure~\ref{fig-tool}, the tool presented two side-by-side panels where users could select and compare any source or target dataset. For each selection, the browser listed all categories with corresponding image counts. This list was sortable alphabetically or by size. Clicking a category revealed a random sample of images that could be refreshed. Crucially, the browser intentionally omitted performance metrics or other metadata to ensure judgments were based solely on visual evidence. This design enabled participants to inspect features like textures and structures and assess class coverage, supporting a relative visual analysis with a low information load.

\newpage
\section{Additional visualization of results}
\begin{figure}[!h]
    \centering
    \includegraphics[width=0.7\linewidth]{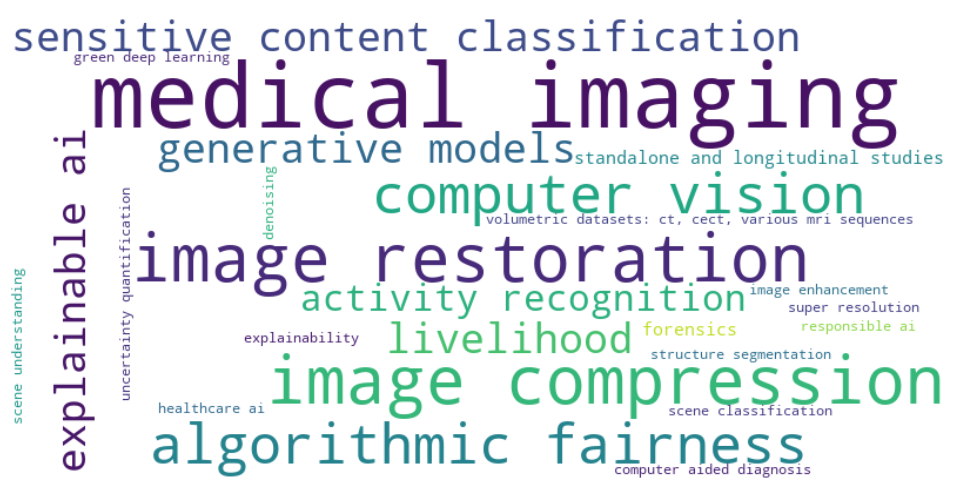}
    \caption{Areas of research expertise among participants.}
    \label{fig:wordcloud}
\end{figure}

\begin{figure}[ht]
    \centering
    \includegraphics[width=\linewidth]{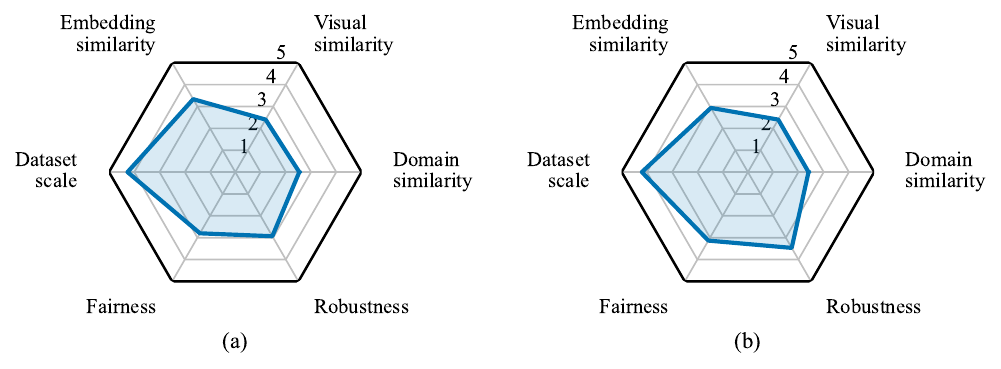}
    \caption{Ratings of expected pretraining effects for a successful fine-tuning outcome presented by a 5-point scale (1 = very poor, 5 = very good). (a) \texttt{CS-tissue}: H\&E patch classification. (b) \texttt{CS-xray}: chest X-ray classification.}
    \label{fig:radar}
\end{figure}

\newpage
\section{Codebook}
Our review of the literature on factors influencing transfer learning (Section \ref{sec:factors}) served as the entry point to coding. Based on these factors, <two anonymized authors> jointly developed a codebook. Each code (N=15) was described through its definition, guidance on when to apply or not apply it, and an example \cite{thompson_guide_2022} (Table~\ref{tab:codebook}). The initial set covered theoretically derived factors while leaving room for emergent codes. The same authors then independently coded all open-ended responses to the case studies (Q19 and Q23), applying the predefined codes and introducing new ones where necessary. They subsequently met to compare their usage of codes, resolve discrepancies, and refine the inductive codes.
\begin{longtable}{p{0.2\linewidth}|p{0.3\linewidth}|p{0.45\linewidth}}
 \toprule
 \textbf{Code} & \textbf{Definition} & \textbf{Examples} \\
 \midrule
 \endfirsthead
 
 \caption*{Codebook for annotating themes (continued)} \\
 \toprule
 \textbf{Code} & \textbf{Definition} & \textbf{Examples} \\
 \midrule
 \endhead
 
 \midrule
 \multicolumn{3}{r}{\textit{continued on next page}} \\
 \endfoot
 
 \bottomrule
 \noalign{\vspace{10pt}} 
 \caption{Codebook for annotating themes in participant answers to the case studies: \texttt{CS-tissue} and \texttt{CS-xray}.}
 \label{tab:codebook} \\
 \endlastfoot
 
 \textbf{researcher experiences} & Widely adopted practice in the community, experience from self or others. & \underline{Positive}: I heard from colleagues and in talks that it works for H\&E images \\ 
 & & \underline{Positive}: recent foundational models trained in TCGA has outperformed the rest of the model \\
 & & \underline{Positive}: based on my experience, the fact that its medical is not always that important \\ 
 \midrule
 \textbf{researcher incentives} & Expectations from the community to use. & \underline{Positive}: reviewers might ask \\ 
 & & \underline{Positive}: must be tested as a baseline \\
 \midrule
 \textbf{source usability} & How quick it is to get started? Worked with it before. & \underline{Negative}:  I never worked with this dataset, so would not select it \\ 
 & & \underline{Positive}: Easy to use \\
 \midrule
 \textbf{source availability} & Pretrained models or data easily available. & \underline{Positive}: Pretrained models are available \\ 
 \midrule
 \textbf{source awareness} & Well-known or popular datasets & \underline{Negative}: Was not aware of it at the time of the research \\
 \midrule
 \textbf{source size} & Refer to the amount of data & \underline{Positive}: As a large-scale dataset in the same radiological domain \\
 \midrule
 \textbf{source diversity} & Describing qualities of the dataset with words like diversity or variability, sometimes not much defined & \underline{Positive}: Large-scale, diverse visual data \\ 
 \midrule
 \textbf{source general purpose} & Refer to general feature extractor, link to robustness and generalization in a good way & \underline{Positive}: Large-scale, diverse visual data that allows models to learn transferable low- and mid-level features \\ 
 & & \underline{Positive}: My experience is that this kind of models are quite OK since they learn useful features.\\
 \midrule
 \textbf{source other evaluations} & Concerns about bias, reliability, could be related to generalization but seems more about not-only-accuracy effects, like bias/fairness & \underline{Negative}: However, they may not be much reliable. \\ 
 \midrule
 \textbf{source quality unspecified} & Mention quality but without definition or context & \underline{Positive}: Good image quality \\
 \midrule
 \textbf{similarity semantic} & Natural images versus medical imaging, also mention specific modalities & \underline{Negative}: I consider that 'natural image' domain dataset would not have a satisfying performance for chest-rays \\
 & & \underline{Positive}: As a large-scale dataset in the same radiological domain \\
 & & \underline{Positive}: Considered because it is a large-scale medical dataset, which may provide more relevant features than natural images \\
 \midrule
 \textbf{similarity visual color} & Visual similarity, difference between black and white and color images & \underline{Positive}: The images are RGB \\
 & & \underline{Positive:} Colour images are usually easier to transfer to other colour images \\
 \midrule
 \textbf{similarity visual texture} & Visual similarity related to texture and shapes & \underline{Positive}:  large part of the image is background \\
 \midrule
 \textbf{similarity unspecified} & Not clear definition of similarity & \underline{Negative}: narrow domain gap from the target domain.\\
\end{longtable}

\section{Full questionnaire} \label{appendix:questionnaire}
\subsection{Private experience}
We’d like to ask a few questions about your background in machine learning and research.

\begin{enumerate}[label=(\arabic*), series=questions]
    \item \textbf{What is your current position?}
    \begin{itemize}[label=\textbullet]
        \item Bachelor student
        \item Master student
        \item PhD student / Doctoral candidate
        \item Postdoctoral researcher
        \item Assistant professor / Lecturer
        \item Associate professor
        \item Full professor
        \item Research assistant
        \item Research scientist / Engineer (non-faculty)
        \item Industry researcher / R\&D engineer
        \item Others
    \end{itemize}

    \item \textbf{How many years of experience in machine learning do you have?}
    \textit{Please include the total number of years you have actively used machine learning methods in your studies, research, or work. This includes coursework, academic projects, publications, or applications in industry.}

    \item \textbf{What is your primary domain or research area (e.g., medical imaging)?}
    \textit{Provide no more than 5 tags, one tag per line / textbox.}

    \item \textbf{What types of transfer learning have you used?}
    \textit{You may choose multiple options or specify your own if it's not listed.}
    \begin{itemize}[label=\checkbox]
        \item Domain adaptation (apply a model to a new domain with different data distribution)
        \item Fine-tuning (start from a pretrained model and update its weights on a new task)
        \item Feature extraction (use a pretrained model to extract features, without updating its weights)
        \item Multi-task learning (train a model on multiple related tasks at the same time)
        \item I have not used transfer learning in a project before
        \item Others: (specify your own)
    \end{itemize}

    \item \textbf{In how many papers have you used transfer learning?}

    \item \textbf{Have you mainly worked with public or private datasets?}
    \begin{itemize}[label=\textbullet]
        \item Mostly public datasets (e.g., ImageNet-1K, COCO)
        \item Mostly private datasets (e.g., proprietary or internal datasets not publicly available)
        \item Both equally
        \item Not sure
    \end{itemize}

    \item \textbf{(Optional) Could you please share the country of your current affiliation with us?}

    \item \textbf{(Optional) If you would be open to a short (around 20-minute) follow-up interview to discuss your answers in more detail, please leave your contact information.}
\end{enumerate}

\subsection{A most recent transfer learning project you’ve worked on}
We would like to ask you a few questions about a project in which you applied transfer learning.

\begin{enumerate}[resume=questions]
    \item \textbf{Which category best describes the project?}
    \textit{You may specify your own if it's not listed.}
    \begin{itemize}[label=\textbullet]
        \item Image classification
        \item Object detection
        \item Semantic segmentation
        \item Natural language processing (e.g., text classification, translation)
        \item Speech processing (e.g., speech recognition, speaker identification)
        \item Time series forecasting or anomaly detection
        \item Medical imaging (e.g., diagnosis, segmentation)
        \item Industrial inspection or quality control
        \item Recommender systems
        \item Cross-modal learning (e.g., image-to-text, text-to-audio)
        \item Few-shot or zero-shot learning
        \item Others: (specify your own)
    \end{itemize}

    \item \textbf{What was the main goal of the project?}
    \textit{You may specify your own if it's not listed.}
    \begin{itemize}[label=\checkbox]
        \item Improve performance on a specific task
        \item Adapt to a new domain
        \item Reduce training time or amount of training data
        \item Improve robustness or generalization
        \item Explore feasibility of transfer learning
        \item Others: (specify your own)
    \end{itemize}

    \item \textbf{What were the \emph{source} and \emph{target datasets}?}
    \textit{Target dataset could also be the one for comparing embeddings if your project does not involve fine-tuning.}

    \item \textbf{What was the model design you use? (e.g., Resnet-50)}

    \item \textbf{What evaluation methods did you use to assess the project?}
    \textit{Examples: F1 score, AUC, feature generalization (e.g., t-SNE), comparison with a baseline without transfer learning, etc. Please list one method per line. You can add more rows if needed. (Max: 8 rows)}

    \item \textbf{What were the reasons for choosing the \emph{source dataset}?} 
    \textit{You may specify your own if it's not listed.}
    \begin{itemize}[label=\checkbox]
        \item Source and target images are visually similar (e.g., texture, shape, etc.)
        \item Source and target images are semantically similar
        \item The amount of data is large enough
        \item I had used it before
        \item It has shown good performance in prior work
        \item It is widely used in the community
        \item It had a pretrained model available
        \item I had a good impression of it
        \item Others:
    \end{itemize}

    \item \textbf{Did you consider other \emph{source datasets}? If yes, why did you not choose them?}
    \begin{itemize}[label=\textbullet]
        \item Yes - \textit{Why did you not choose them?}
        \item No
    \end{itemize}
    
\end{enumerate}

\subsection{Case studies}
\subsubsection{Case study 1: \texttt{CS-tissue}}
In this task, we aim to develop a transfer-learning pipeline for nine-class patch-level tissue classification in colorectal Hematoxylin and Eosin (H\&E) images. A large source model trained on the selected \emph{source dataset} will be fine-tuned on a lean subset of the CRC-VAL-HE-7K target set, then evaluated on the remaining, unseen patches to verify generalization across new patients and subtle staining shifts. Below are the summary of the target and \emph{source datasets}:

\vspace{1em}
\noindent\textbf{Target dataset}: CRC-VAL-HE-7K \\
\textbf{Size \& granularity}: 7,180 non-overlapping H\&E patches, each $224\times224$ at 0.5 $\mu$m/pixel. \\
\textbf{Patients}: 50 individuals with colorectal adenocarcinoma. \\
\textbf{Classes}: Adipose (ADI), Background (BACK), Debris (DEB), Lymphocytes (LYM), Mucus (MUC), Smooth-muscle (MUS), Normal Mucosa (NORM), Stroma (STR), Tumour Epithelium (TUM). \\
\textbf{Dataset split}: Randomly sample 250 patches per class for training / validation; all remaining patches (patient-disjoint from training) for testing. \\
\textbf{Performance criteria}: Macro-AUC.
\vspace{1em}

\begin{table*}[!ht]
    \centering
    \begin{tabularx}{\textwidth}{@{} l XXX @{}}
        \toprule
        \textbf{Feature} & \textbf{ImageNet-1K} & \textbf{RadImageNet} & \textbf{Ecoset} \\
        \midrule
        Primary Content & General everyday objects and fine-grained concepts (e.g., animals, instruments, plants, structures). & Radiological images (CT, MRI, Ultrasound) across various pathologies and anatomies (e.g., lung, brain, liver). & Everyday objects and coarse concepts selected based on linguistic frequency and human relevance. \\
        \midrule
        Number of Images & $\approx$ 1.3 million training images and 50,000 validation images. & $\approx$ 1.35 million annotated images. & Over 1.5 million images. \\
        \midrule
        Number of Classes & 1,000 object classes. & 165 distinct pathologies. & 565 basic-level categories. \\
        \midrule
        Primary Use Case & Benchmarking general-purpose computer vision models for tasks like image classification and object detection. & Transfer learning and developing specialized deep-learning models for medical image analysis. & Training and testing models to better align with human vision and object-recognition behavior. \\
        \midrule
        Key Distinction & Serves as a de facto standard for pretraining models and comparing algorithm performance. & Domain-specific dataset intended to improve model performance on medical tasks compared to models pretrained on non-medical data like ImageNet-1K. & Created to be more representative of objects relevant to humans than ImageNet-1K, with a focus on concrete categories. \\
        \bottomrule
    \end{tabularx}
\end{table*}

\vspace{2em}

\begin{enumerate}[resume=questions]
    \item \textbf{How likely would you consider the following datasets as the source for this task? You may also specify your own if it's not listed.}
    \begin{center}
        \begin{tabular}{lcccc}
        \toprule
        & \textbf{Likely} & \textbf{Neutral} & \textbf{Unlikely} & \textbf{Not sure} \\
        \midrule
        ImageNet-1K & $\bigcirc$ & $\bigcirc$ & $\bigcirc$ & $\bigcirc$ \\
        RadImageNet & $\bigcirc$ & $\bigcirc$ & $\bigcirc$ & $\bigcirc$ \\
        Ecoset & $\bigcirc$ & $\bigcirc$ & $\bigcirc$ & $\bigcirc$ \\
        Your suggested dataset: & $\bigcirc$ & $\bigcirc$ & $\bigcirc$ & $\bigcirc$ \\
        \bottomrule
        \end{tabular}
    \end{center}

    \vspace{1em}
    \item \textbf{How would you subjectively assess the expected fine-tuning performance on each of the following datasets?}
    \begin{center}
        \begin{tabular}{lccccc}
            \toprule
            & \textbf{\begin{tabular}{@{}c@{}}Very\\poor\end{tabular}} & \textbf{Poor} & \textbf{Moderate} & \textbf{Good} & \textbf{\begin{tabular}{@{}c@{}}Very\\good\end{tabular}} \\
            \midrule
            ImageNet-1K & $\bigcirc$ & $\bigcirc$ & $\bigcirc$ & $\bigcirc$ & $\bigcirc$ \\
            RadImageNet & $\bigcirc$ & $\bigcirc$ & $\bigcirc$ & $\bigcirc$ & $\bigcirc$ \\
            Ecoset & $\bigcirc$ & $\bigcirc$ & $\bigcirc$ & $\bigcirc$ & $\bigcirc$ \\
            Your suggested dataset: & $\bigcirc$ & $\bigcirc$ & $\bigcirc$ & $\bigcirc$ & $\bigcirc$ \\
            \bottomrule
        \end{tabular}
    \end{center}
    \vspace{1em}

    \item \textbf{How would you rate the expected effect of pretraining on each \emph{source dataset}, after fine-tuning on the target task?}
    \textit{Please assess the model you will obtain, not the datasets themselves. You may specify your own criteria if it's not listed.}
    
    Participants were asked to provide a rating for each cell based on the scale: \textbf{Very poor, Poor, Moderate, Good, Very good.}

    \vspace{1em}
    \begin{center}
        \begin{tabular}{@{} >{\raggedright\arraybackslash}p{6cm} c c c c @{}}
            \toprule
            & \textbf{ImageNet-1K} & \textbf{RadImageNet} & \textbf{Ecoset} & \textbf{Your dataset} \\
            \midrule
            Domain similarity (e.g., semantic content aligns with target task) & & & & \\
            Visual similarity (e.g., visual resemblance) & & & & \\
            Embedding similarity (i.e., the extracted feature representation) & & & & \\
            Dataset scale (i.e., sample size, number of classes) & & & & \\
            Fairness (e.g., demographic bias) & & & & \\
            Robustness (e.g., noise, domain shift, imbalance) & & & & \\
            Your suggested criteria: & & & & \\
            \bottomrule
        \end{tabular}
    \end{center}

    \vspace{1em}
    \item \textbf{Why did you consider or did not consider each dataset as a suitable source for this task?}
\end{enumerate}

\subsubsection{Case study 2: \texttt{CS-xray}}
In this task we aim to develop a transfer-learning pipeline for multi-label chest X-ray classification. Starting from a model trained on the selected \emph{source dataset}, we will fine-tune it on a small subset from the CheXpert dataset, then evaluate how well it detects common thoracic pathologies when only a small, label-balanced slice of the target data is available for fine-tuning. To focus on labels that are well represented, all categories with fewer than 100 cases were dropped. Below are the summary of the target and \emph{source datasets}:

\vspace{1em}
\noindent\textbf{Target dataset}: CheXpert \\
\textbf{Size \& granularity}: 834 anterior-posterior, posterior-anterior, and lateral CXRs (typically down-sampled to $320\times320$). \\
\textbf{Patients}: 662 unique patients (one study per patient). \\
\textbf{Classes}: Only labels with $\ge 100$ images are retained: Atelectasis, Cardiomegaly, Edema, Enlarged Cardiomediastinum, Lung Opacity, No Finding, Pleural Effusion, Support Devices. The sparse labels Consolidation, Fracture, Lung Lesion, Pleural Other, Pneumonia, and Pneumothorax are removed. All labels were annotated and verified by human experts. \\
\textbf{Dataset split}: Randomly sample 50 images per retained label for training / validation; all remaining images ($\sim$430+) from the other studies (patient-disjoint from training) for testing. \\
\textbf{Performance criteria}: Macro-AUC.

\vspace{2em}
\textit{For this case study, participants were asked the same set of questions (Questions 16-19) regarding the same \emph{source datasets} as in Case Study 1 (\texttt{CS-tissue)}.}


\end{document}